\documentclass[9pt,a4paper]{rho-class/rho}
\usepackage[english]{babel}
\usepackage{bm}
\usepackage{booktabs}
\usepackage{comment}
\usepackage{tabularx}
\usepackage{algorithm}
\usepackage{algpseudocode}
\usepackage{multicol}
\usepackage{caption}
\usepackage{float}

\setbool{rho-abstract}{true} 
\setbool{corres-info}{true} 

\title{Evaluating Rare Disease Diagnostic Performance in Symptom Checkers: A Synthetic Vignette Simulation Approach}

\author[1,$\dagger$]{Takashi Nishibayashi}
\author[1]{Seiji Kanazawa, MD, PhD}
\author[1]{Kumpei Yamada}

\affil[1]{Ubie, Inc.}

\dates{This manuscript was compile on June 24, 2025}

\leadauthor{N.Takashi et al.}
\theday{June 24, 2025} 

\corres{Takashi Nishibayashi}
\email{takashi.nishibayashi@dr-ubie.com}

\license{}

\newcommand{\abssection}[1]{%
  \vspace{0.5em}
  \noindent\textbf{#1:}
}
\newcommand{\demographics}{\mathbf{u}}
\newcommand{\responses}{\mathbf{r}}
\newcommand{\potentialresponses}{\mathbf{r}_\mathrm{potential}}
\newcommand{\collectedresponses}{\mathbf{r}_\mathrm{collected}}
\newcommand{\allsymptoms}{\mathcal{S}}
\newcommand{\preddiseases}{\bm{\hat d}}
\newcommand{\ithpreddiseases}{\bm{\hat d}_i}
\newcommand{\truedisease}{d_\mathrm{true}}
\newcommand{\ithtruedisease}{d_{\mathrm{true},i}}
\newcommand{\allcauses}{\mathcal{C}}
\newcommand{\rarediseases}{\mathcal{C}^\mathrm{rare}}
\newcommand{\allphenotype}{\mathcal{P}}
\newcommand{\dataset}{\mathcal{D}}
\newcommand{\datasetrare}{\mathcal{D}^\mathrm{rare}}
\newcommand{\datasetcommon}{\mathcal{D}^\mathrm{common}}
\newcommand{\estimatedimpact}{\Delta {M}_d}

\newcommand{\annotationofd}{\mathcal{A}_{d}}

\begin{abstract}
\abssection{Background} Symptom Checkers (SCs) provide medical information tailored to user symptoms. A critical challenge in SC development is preventing unexpected performance degradation for individual diseases, especially rare diseases, when updating algorithms. This risk stems from the lack of practical pre-deployment evaluation methods. For rare diseases, obtaining sufficient evaluation data from user feedback is difficult, and creating numerous clinical vignettes is costly and impractical.

\abssection{Objective} To evaluate the impact of algorithm updates on the diagnostic performance for individual rare diseases before deployment, this study proposes and validates a novel Synthetic Vignette Simulation Approach. This approach aims to enable this essential evaluation efficiently and at a low cost.

\abssection{Methods} To estimate the impact of algorithm updates, we generated synthetic vignettes from disease-phenotype annotations in the Human Phenotype Ontology (HPO), a publicly available knowledge base for rare diseases curated by experts. Using these vignettes, we simulated SC interviews to predict changes in diagnostic performance. The effectiveness of this approach was validated retrospectively by comparing the predicted changes with actual performance metrics (reproduced in offline tests) using the R-squared ($R^2$) coefficient.

\abssection{Results} Our experiment, covering eight past algorithm updates for rare diseases, showed that the proposed method accurately predicted performance changes for diseases with phenotype frequency information in HPO (n=5). For these updates, we found a strong correlation for both Recall@8 change ($R^2=0.83$,$p=0.031$) and Precision@8 change ($R^2=0.78$,$p=0.047$). In contrast, updates for diseases without frequency data (n=3) resulted in large prediction errors, highlighting the critical role of this information. Mapping HPO phenotypes to SC symptoms required approximately 2 hours per disease.

\abssection{Conclusions} Our proposed method enables the pre-deployment evaluation of SC algorithm changes for individual rare diseases. This evaluation is based on a publicly available medical knowledge database created by experts, ensuring transparency and explainability for stakeholders. 
Additionally, SC developers can efficiently improve diagnostic performance for rare diseases at a low cost.
\end{abstract}


\begin{document}
	
    \maketitle
    \thispagestyle{firststyle}
    \nolinenumbers

\begin{multicols}{2}
\section{Introduction}

\subsection{Background: Symptom Checkers and the Need for Performance Evaluation}
Accessing health information via the Internet has become common and is now considered the first step in seeking medical care [\cite{Bachl2024-gw,Van_Kessel2023-qi}]. Symptom Checkers (SCs), available as websites or mobile applications, provide personalized medical information based on users' demographic data and symptoms. SCs are expected to alleviate the burden on local medical resources amid a global shortage of healthcare professionals and guide patients to appropriate medical care when necessary [\cite{Berry2018-xr,McIntyre2020-vg,Gottliebsen2020-vn}]. Some government health agencies have officially adopted and promoted these tools [\cite{NHS-symptomchecker,covid-symptomchecker}].

Some SCs support the diagnosis of rare and genetic diseases [\cite{Fujiwara2018-yk}]. 
Even though delays in treatment can be fatal, 95\% of patients do not receive appropriate care because of the difficulty in diagnosing these diseases, which takes an average of 5-7 years in the US and UK [\cite{rarediseasefact-sc,Office_of_the_Commissioner2022-oh,Ifpma2017-cx}]. SCs are expected to identify potential rare diseases early based on patient signs and symptoms, guiding them to suitable medical institutions promptly.

With the widespread use of SCs, there is a growing demand for objectively evaluating their diagnostic performance. The effectiveness of SCs primarily depends on their diagnostic accuracy. While SCs utilize AI or algorithms to suggest potential diagnoses, incorrect recommendations can prevent users from receiving appropriate medical care. The objective evaluation of SCs mainly focuses on the validity of their diagnosis and triage [\cite{Riboli-Sasco2023-ph,Wallace2022-mq}]. Clinical vignettes, fictional medical histories created by physicians, are commonly used for evaluation. Although there are limitations to using clinical vignettes [\cite{El-Osta2022-rz}], they provide general and objective results, making them a standard method. Studies compare SC outputs with physician's diagnoses [\cite{Peven2023-su}], evaluate the performance of multiple SCs [\cite{Ceney2021-gi}], and report self-evaluations by SC providers [\cite{Baker2020-bw,Hammoud2024-hz,Richens2020-fs}].

\subsection{Challenges in Evaluating Symptom Checker Performance for Rare Diseases: A Developer's Perspective}
SC developers continuously work on improving diagnostic performance, including for individual rare diseases. Each time they update the algorithm, they want to evaluate performance for specific diseases objectively. This evaluation ensures improvements in the target diseases and that diagnostic accuracy for other diseases does not decrease. Without sufficient performance evaluation, algorithm updates that degrade performance and go unnoticed can be released. For rare diseases, there is no established method to mitigate this risk. Generally, SC performance testing uses vignettes, but creating sufficient clinical vignettes for each disease is costly and impractical. Few specialists are capable of making these vignettes for rare diseases. 
Developers need more than 100 vignettes per disease to achieve a 95\% confidence interval for sensitivity smaller than 0.1. The vignette set used in Hammoud et al.'s study [\cite{Hammoud2024-hz}] covers 254 diseases with 400 vignettes in total, having a maximum of 5 vignettes per disease, which is insufficient for evaluating disease-specific performance. Some studies construct performance evaluation datasets based on case reports [\cite{Miyachi2023-sr}], but case reports often cover rare instances and do not adequately represent typical cases. An ideal evaluation dataset would comprehensively cover typical cases across various age groups and genders, including early and late stages of disease onset. While SC developers can collect user feedback continuously to evaluate performance, the limited feedback for rare diseases makes data collection time-consuming. 

\subsection{Proposal: Simulating Real-World Symptom Checker Performance with Medical Knowledge Databases}
To address this issue, we propose a method for evaluating SC diagnostic performance for rare diseases using a medical knowledge database. A single medical professional cannot gain extensive clinical experience in rare diseases. This limitation has driven the development of medical knowledge databases to facilitate knowledge sharing and computational analysis. Orphanet [\cite{Weinreich2008-mi}] and OMIM [\cite{Hamosh2005-op}] are well-known, and a database of disease phenotype annotations coded using the Human Phenotype Ontology (HPO) [\cite{Gargano2024-fc}], which systematizes abnormal clinical phenotypes in humans, is available. Disease phenotype annotations consist of clinical phenotypes that occur in diseases and their frequencies. We intended that synthetic vignettes generated from medical knowledge databases could serve as inputs for SCs. By simulating interviews with these vignettes, we aimed to estimate the impact of algorithm updates on service performance. Our proposed method has the following features:
\begin{itemize}
\item Transparency and Explainability: Our evaluation relies on publicly available and peer-reviewed medical knowledge databases curated by experts. This foundation provides a high degree of transparency and explainability to stakeholders.
\item Cost-effectiveness: Synthetic vignettes are significantly more cost-effective to generate compared to traditional clinical vignettes.
\item Adaptability and Developer Friendliness: Our method is not dependent on the specific AI or algorithms employed by an SC, making it adaptable for various SC developers. This adaptability ensures robustness to internal implementation changes, enhancing developer friendliness.
\end{itemize}
An R-squared coefficient of determination between the estimated values and actual performance changes of 0.7 or higher would provide reference values for developers. To demonstrate the effectiveness of the proposed method, we verified that it is possible to evaluate the impact of updating the algorithm on Ubie Symptom Checker [\cite{ubie}] and published the results.

\section{Method}

\subsection{Overview}

\begin{figure*}[t!]
    \centering
    \includegraphics[width=0.95\linewidth]{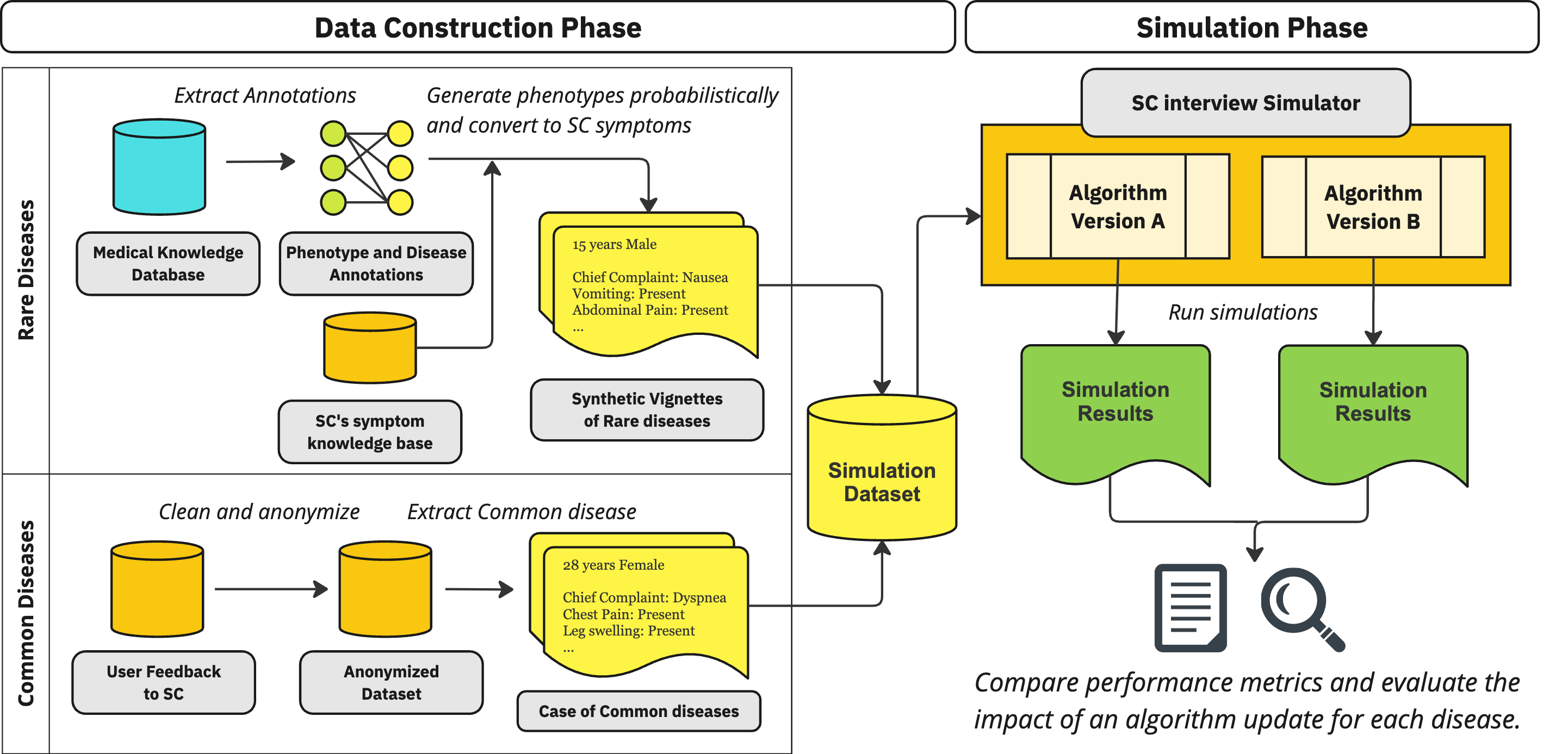}
    \caption{Overall flowchart of our proposal method.}
    \label{fig:overview}
\end{figure*}

In this chapter, we first give an overview of the proposed method. Next, we describe the details of the proposed method, including the details of the data that appears in each phase and the formulation of the processing. Figure \ref{fig:overview} shows an overview of our proposed method. Our method consists of two phases. The first phase is the construction of the dataset. The second phase is the simulation of the interview and estimating the impact using the constructed dataset. The first phase consists of the generation of vignettes for rare diseases and the generation for common diseases.

\subsection{Problem Formulation}
We clarify the input and output of the SCs' symptom check process to construct the data set and derive diagnostic performance metrics. The SC user first declares their attributes, such as age, gender, race, and chief complaint, and begins the interview process. In the following dialogue phase, the patient interacts with the system to answer questions about additional symptoms and medical history. Finally, the system ends the interview session and outputs several potential causes based on the information gathered.

The information obtained from the patient can be divided into two types and characterized using a tuple $(\demographics,\responses)$. Let $\demographics$ (User Demographics) be a set of individual patient attributes such as age and gender.
Each $\demographics$ has individual patient attributes such as age or gender $u_i$ as $\demographics = \{u_1, u_2,\ldots,u_m\}$.
Let $\responses$ (Responses) be the information obtained for each symptom by the system through dialogue with the patient, in addition to the symptoms explicitly given by the patient.
Let $\allsymptoms$ be the set of all symptoms the SC system can handle. Let $a$ be the patient's response to a symptom $s \in \allsymptoms$, where $a \in \{\rm{present}, \rm{absent}, \rm{unknown}\}$.
The information obtained from one question is a tuple $(s, a)$.
Let $\potentialresponses$ be the set of all possible responses to a patient's condition, where $\potentialresponses = \{(s_1, a_1), (s_2, a_2), \ldots, (s_p,a_p)\}$ that can be given based on the patient's condition.
We use $\collectedresponses$ to denote the information collected by the system through dialogue with the patient, then $\collectedresponses$ is a subset of $\potentialresponses$. Namely, $\collectedresponses \subseteq \potentialresponses$.

Let $\allcauses$ be the set of all potential causes that the system can output. $\allcauses$ is mainly composed of disease names, but it also includes items that are not diseases. The system ranks the top-K potential causes $d \in \allcauses$ based on the information obtained from the patient $(\demographics, \collectedresponses)$ and presents the top-K potential causes.
Top-K potential causes is a set $\preddiseases = \{d_1, d_2,\ldots,d_K : d \in \allcauses\}$.
We use $f$ to denote the symptom checker algorithm, including the process of interacting with the patient. Namely
\begin{equation} \label{ec:equation}
f:\demographics, \potentialresponses \rightarrow \preddiseases.
\end{equation}


\subsection{Simulation Dataset}

Let $\dataset$ be an evaluation dataset for measuring the diagnostic performance of SC by simulating the symptom check process. $\dataset$ contains multiple medical interview sessions. For each medical interview session, the correct potential cause label $\truedisease \in \allcauses$ is given as the gold standard. The data for simulating each interview session is $(\demographics, \potentialresponses, \truedisease)$. Let $\rarediseases$ be an rare disease group of potential causes. Namely, $\rarediseases \subseteq \allcauses$. Define the sub-dataset of the dataset $\dataset$ using $\datasetrare$ and $\datasetcommon$
\begin{align} \label{ec:1}
\datasetrare &= \{(\demographics, \potentialresponses, \truedisease) \in \dataset: \truedisease \in \rarediseases\} \\
\datasetcommon &= \{(\demographics, \potentialresponses, \truedisease) \in \dataset: \truedisease \notin \rarediseases\}
\end{align}

\subsection{Simulation and Evaluation Metrics of Diagnostic Accuracy}
The right side of Figure \ref{fig:overview} shows how to perform simulations for each of the two versions of the SC algorithm and compare the results. For each row of the dataset $\dataset$, the algorithm $f$ first receives the patient demographics and chief complaints, then actively collects patient information, and finally returns multiple potential causes. Let $n$ be the size of the dataset, and let $\ithpreddiseases, \ithtruedisease$ be the output and true label of the $i$-th simulation, respectively. We adopt the Top-K Recall(Recall@K) and Top-K Precision(Precision@K), defined below, as performance evaluation metrics for each potential cause.

\begin{align}
\text{Top-K Recall of}\ d &= \frac{\sum_{i=1}^n{\mathbb{1}[\ \ithtruedisease \in \ithpreddiseases \land \ithtruedisease = d\ ]}}{\sum_{i=1}^n{\mathbb{1}[\ \ithtruedisease = d\ ]}} \\[4pt]
&= \frac{\text{The \# of true positive of }d}{\text{The \# of simulations } \truedisease\text{ is }d}
\end{align}

\begin{align}
\text{Top-K Precision of}\ &d = \frac{\sum_{i=1}^n{\mathbb{1}[\ \ithtruedisease \in \ithpreddiseases \land \ithtruedisease = d\ ]}}{\sum_{i=1}^n{\mathbb{1}[\ d \in  \ithpreddiseases\ ]}} \\[4pt]
&= \frac{\text{The \# of true positive of }d}{\text{The \# of simulations }d\text{ is predicted}}
\end{align}

Both are widely used as performance evaluation metrics for recommender systems, formulated as ranking problems. Recall is the same as Sensitivity, and Precision is the same as Positive Predictive Value.
\subsection{Estimating the impact of algorithm enhancement for specific disease}

Let $\mathcal{M} \in \{\mathcal{M}_{\rm recall}, \mathcal{M}_{\rm precision}\}$ denote the simulation process to calculate metrics of $d$ as follows

\begin{equation}
\mathcal{M}:f, \dataset, d \rightarrow [0, 1].
\end{equation}

Using the algorithm $f_{\rm before}$ before the update and the algorithm $f_{\rm after}$ after the update for the disease $d$, we can calculate the estimated impact of the algorithm update on disease d as follows

\begin{equation}
\estimatedimpact = \mathcal{M}(f_{\rm after}, \mathcal{D},d) - \mathcal{M}(f_{\rm before}, \mathcal{D},d).
\end{equation}

\subsection{Dataset construction of rare diseases}

Our proposed method constructs a dataset $\datasetrare$ based on the HPO phenotypic annotations (HPO annotations) for simulating an interview session of patients with rare diseases.
Some HPO Annotations are shown in Table \ref{table:hpoannotation}. HPO Annotations list the HPO phenotypes associated with each disease and their frequencies.
Let $\allphenotype$ (phenotypes) be the set of all phenotypes defined by HPO. Let $p \in \allphenotype$ be a phenotype. Let $q$ be a frequency of expression defined by HPO: $q \in \{{\rm very\ frequent}, {\rm frequent},\ldots,{\rm not\ specified}\}$.
Table \ref{table:freqs} shows the frequencies of expression defined by HPO.
We denote the set HPO annotations for disease $d$ as $\annotationofd = \{(p_1,q_1),(p_2,q_2),\ldots,(p_k,q_k)\}$.

\begin{table}[H]
    \RaggedRight
    \caption{Part of the HPO Annotations.}
    \label{table:hpoannotation}
    \begin{tabularx}{\linewidth}{lll}
    \toprule
    Disease & Phenotype & Frequency \\
    \midrule
    Blau syndrome & Facial palsy & Occasional \\
     & Xerostomia & Occasional \\
     & Glaucoma & Frequent \\
     & Camptodactyly of finger & Frequent \\
     & Joint swelling & Very Frequent \\
     & Erythema & Very Frequent \\
    \bottomrule
    \end{tabularx}
        
    \tabletext{Human Phenotype Ontology version 2024-02-08.}
\end{table}

\begin{table}[H]
  \centering
  \caption{Examples of HPO Frequency Options and Definitions.}
  \begin{tabularx}{\linewidth}{lll}
    \toprule
    Label or Value & Definition \\
    \midrule
    Very frequent & Present in 80\% to 99\% of the cases. \\
    Frequent & Present in 30\% to 79\% of the cases. \\
    Occasional & Present in 5\% to 29\% of the cases. \\
    Very rare & Present in 1\% to 4\% of the cases. \\
    12/45 & A count of patients affected within a cohort. \\
    17\% & A percentage value. \\ 
    <blank> & Not Specified \\
    \bottomrule
  \end{tabularx}
  \label{table:freqs}
\end{table}

SC interview simulator cannot use the HPO phenotype directly because the simulation is performed on SC symptoms.
Therefore, to generate simulation data $\potentialresponses$ from HPO Annotations, we convert the HPO phenotype into SC symptoms.
This process is as follows,
\begin{equation}
g: p \rightarrow \bold{s}
\end{equation}
where $\bold{s} \subset \allsymptoms$.
Multiple mappings exist for HPO phenotypes because the symptoms handled by HPO phenotypes and SCs do not always map one-to-one.
For example, we can map HPO's Acroparesthesia to SC symptoms such as "pain in the fingers," "numbness in the fingers," and more. 
There is a method for incorporating external ontologies into the SC knowledge base [\cite{Stoilos2018-yn}].
Similarly, there are cases where the granularity of diseases handled by SC does not match the diseases in external knowledge bases. 
Table \ref{table:mappings} shows examples of the mapping between HPO phenotypes and SC symptoms.

\begin{table}[H]
  \centering
  \caption{Example of HPO Phenotype to SC Symptoms mapping.}
  \begin{tabularx}{\linewidth}{l X}
    \toprule
    HPO Phenotype & SC Symptom \\
    \midrule
    Purpura & Subcutaneous hemorrhage \\
    Hyperthyroidism & Heat intolerance, Tachycardia, Palpitation, Fine tremor of the fingers, Easy fatigability, Weight loss \\
    Acroparesthesia & Finger pain, Palm pain, Dorsal foot pain, Sole pain, Heel pain, Toe pain, Finger numbness, Toe numbness, Superficial sensory loss, Paresthesia, Burning sensation in the hands and feet \\
    \bottomrule
  \end{tabularx}
  \label{table:mappings}
\end{table}

As you can see, the HPO Annotations do not include patient demographic information. As we could not find a structured database of prevalence and gender ratios by age group, we prepared a distribution of patient demographic information for each disease, with reference to Orphanet [\cite{Weinreich2008-mi}] and other sources, to compensate.
Algorithm \ref{alg:1} shows the procedure for constructing dataset $\datasetrare$.
We call each record of the dataset a synthetic vignette.
The algorithm generates phenotypes based on the manifestation probabilities specified in the HPO Annotations and adds the corresponding SC symptoms to the synthetic vignette. For annotations without specified frequencies, we use a fixed value for the frequency.

\begin{algorithm}[H]
    \caption{Generate simulation dataset from HPO Annotations}
    \label{alg:1}
    \begin{algorithmic}[1]
        \algnewcommand{\LineComment}[1]{\State \(\triangleright\) #1}
        \renewcommand{\algorithmicrequire}{\textbf{Input:}}
        \renewcommand{\algorithmicensure}{\textbf{Output:}}
        \let\oldReturn\Return
        \renewcommand{\Return}{\State\oldReturn}
        \algnewcommand{\Initialize}[1]{%
          \State \textbf{Initialize:}
          \Statex \hspace*{\algorithmicindent}\parbox[t]{.8\linewidth}{\raggedright #1}
        }
        \algnewcommand{\Inputs}[1]{%
          \State \textbf{Inputs:}
          \Statex \hspace*{\algorithmicindent}\parbox[t]{.8\linewidth}{\raggedright #1}
        }
        \algnewcommand{\Outputs}[1]{%
          \State \textbf{Outputs:}
          \Statex \hspace*{\algorithmicindent}\parbox[t]{.8\linewidth}{\raggedright #1}
        }
        
        \Inputs {Target diseases $\bm{d}_{\rm targets} \subset \rarediseases$, \\ The number of generates for each disease $n > 0$}
        \Outputs {Dataset $\datasetrare$}
        \Initialize{$\datasetrare \gets \emptyset$}
        \ForAll{$d \in \bm{d}_{\rm targets}$}
            \For{$i \gets 1$ to $n$}
                \State ${\potentialresponses \gets \emptyset}$
                \ForAll{$(p, q) \in \annotationofd$}
                    \State $v \gets \text{random value between 0 and 1}$
                    \LineComment{Turn on the phenotype $p$ stochastically.}
                    \If {$v < q$} 
                        \State ${\bm s} \gets g(p)$
                        \ForAll{$s \in \bm{s}$}
                            \State $a \gets \text{"present"}$ 
                            \State Add $(s, a)$ to $\potentialresponses$
                        \EndFor
                    \EndIf
                \EndFor
                \State $\demographics \gets$ sample from demographic distribution of $d$
                \State Add $(\demographics, \potentialresponses, d)$ to $\datasetrare$
            \EndFor
        \EndFor
    \end{algorithmic}
\end{algorithm}

\subsection{Dataset construction of common diseases}
SC providers can collect user feedback while operating the service and build datasets. Because user feedback contains many common diseases, anonymizing it makes it possible to use it as a usable dataset.

\subsection{Experiments}
\subsubsection{Evaluate the effectiveness by retrospective study}
We conducted a retrospective study using data obtained from Ubie SC to evaluate the effectiveness of the proposed method. 
We used user feedback received with the consent of users for research use in Ubie SC. We experimented to see if the proposed method could estimate the impact of past algorithm updates for rare diseases in Ubie SC. We used the algorithm before and after the update for a rare disease $d$ to calculate the impact of the algorithm update, Top-8 $\Delta{M}_d$. 
We reproduced the impact of past algorithm updates using user feedback in offline testing. 
We compared the estimated values of the proposed method, $\Delta M_d$, with the actual change reproduced in the offline testing to check whether the two were correlated.

\subsubsection{Setup simulation dataset of rare diseases}
We found 20 diseases for which Ubie SC is predictable and for which HPO Annotation exists.
We selected eight diseases with more than 30 user feedback for calculating actual performance metrics, for which there was a history of algorithm update, and included them in the experiment.
Table \ref{table:targetdiseases} lists the target diseases and an overview of the HPO annotations. There is no frequency information for some diseases, so we have indicated whether or not frequency information is available.
We constructed the dataset $\datasetrare$ for these diseases. We generated 100 cases per disease.
A software engineer (TN) mapped the HPO phenotypes to the symptoms of Ubie SC. A physician (SK) reviewed the results.
At that time, we measured the time taken for each disease.

\begin{table*}[ht]
  \centering
  \captionsetup{font=large}
  \caption{Diseases included in experiments and HPO$\rm{^a}$ annotation summary.}
  \begin{tabularx}{\linewidth}{lllcc}
    \toprule
    Name & Disease ID & Source Database & \# of annotations & Frequency annotation$\rm{^b}$\\
    \midrule
    Behcet's disease (BD) & ORPHA:117 & Orphanet & 68 & \checkmark \\
    Neuromyelitis optica spectrum disorder (NMOSD) & ORPHA:71211 & Orphanet & 12 & \checkmark \\
    Scleroderma & ORPHA:90291 & Orphanet & 65 & \checkmark \\
    Myasthenia gravis & ORPHA:589 & Orphanet & 29 & \checkmark \\
    Systemic lupus erythematosus (SLE) & ORPHA:536 & Orphanet & 36 & \checkmark \\
    Multiple sclerosis & OMIM:126200 & OMIM & 12 & \ding{55} \\
    Crohn's disease & OMIM:266600 & OMIM & 11 & \ding{55} \\
    Parkinson's disease (PD) & OMIM:168600 & OMIM & 29 & \ding{55} \\
    \bottomrule
  \end{tabularx}
  \label{table:targetdiseases}
  \vspace{2pt}
  \begin{minipage}{\linewidth}
    \begin{flushleft}
      $\rm{^a}$HPO: Human Phenotype Ontology. \\
      $\rm{^b}$Frequency annotation: Whether or not the frequency is set for each record in the HPO Annotation.
    \end{flushleft}
  \end{minipage}
  
\end{table*}

\subsubsection{Setup simulation dataset of common diseases}
We constructed $\datasetcommon$ based on user feedback for Ubie SC, except for diseases included in the experiment. The total number of cases in $\datasetcommon$ is 16,715.

\section{Results}

We show our main results in Table \ref{table:mainresult}. It shows the estimated metric change of the target disease derived by our proposed method for each algorithm update and the actual change. The performance metrics were Recall@8 and Precision@8. Actual changes were reproduced in the offline test. For example, our method predicted that SLE's recall would decrease by 0.258 in the algorithm update, which actually decreased by 0.373. Similarly, for the update for scleroderma, our method predicted that the scleroderma's recall would increase by 0.330; it actually increased by 0.147. For the update for Parkinson's disease, our method predicted that the precision would increase by 0.045 but actually decreased by 0.126.

\begin{figure*}[ht!]
    \centering
    \captionsetup{aboveskip=6pt,font=large}
    \begin{subfigure}[b]{0.45\linewidth} 
        \includegraphics[width=\linewidth]{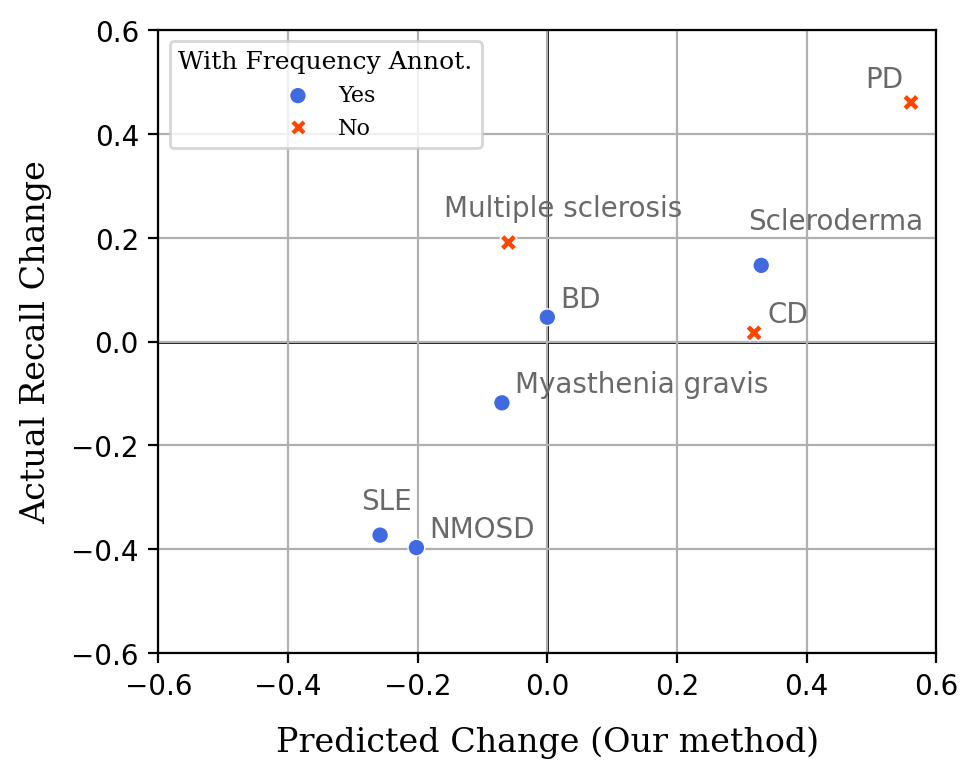}
        \captionsetup{aboveskip=4pt}
        \caption{Recall@8}
        \label{fig:all_recall}
    \end{subfigure}
    \hspace{12pt}   
    \begin{subfigure}[b]{0.45\linewidth} 
        \centering
        \includegraphics[width=\linewidth]{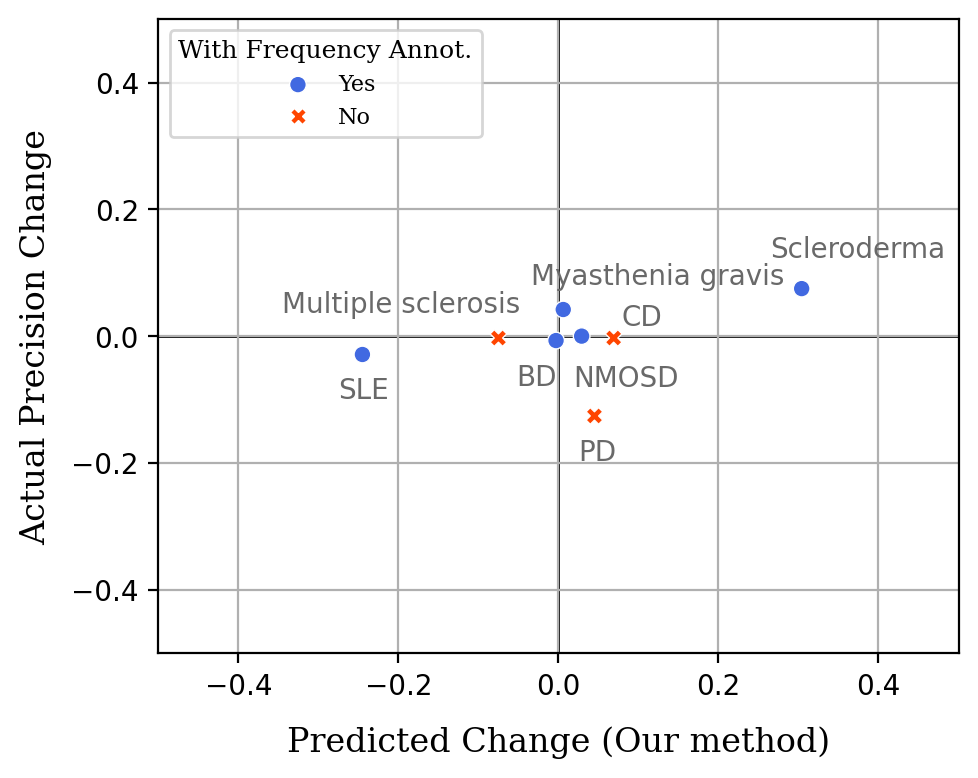}
        \captionsetup{aboveskip=4pt}
        \caption{Precision@8}
        \label{fig:all_precision}
    \end{subfigure}
    \caption{Correlation between predicted and actual change in diagnostic performance. \\ All diseases added to the experiment.}
    \label{fig:plot_all}
\end{figure*}

Figure \ref{fig:plot_all} illustrates the correlation between the estimated changes by our method and the actual changes. Each data point corresponds to each row in Table \ref{table:mainresult}. We divide markers into those with and without frequency information in the HPO annotation, and the differences between the two are visualized.

\begin{figure*}[ht!]
    \centering
    \captionsetup{aboveskip=6pt,font=large}
    \begin{subfigure}[b]{0.45\linewidth} 
        \includegraphics[width=\linewidth]{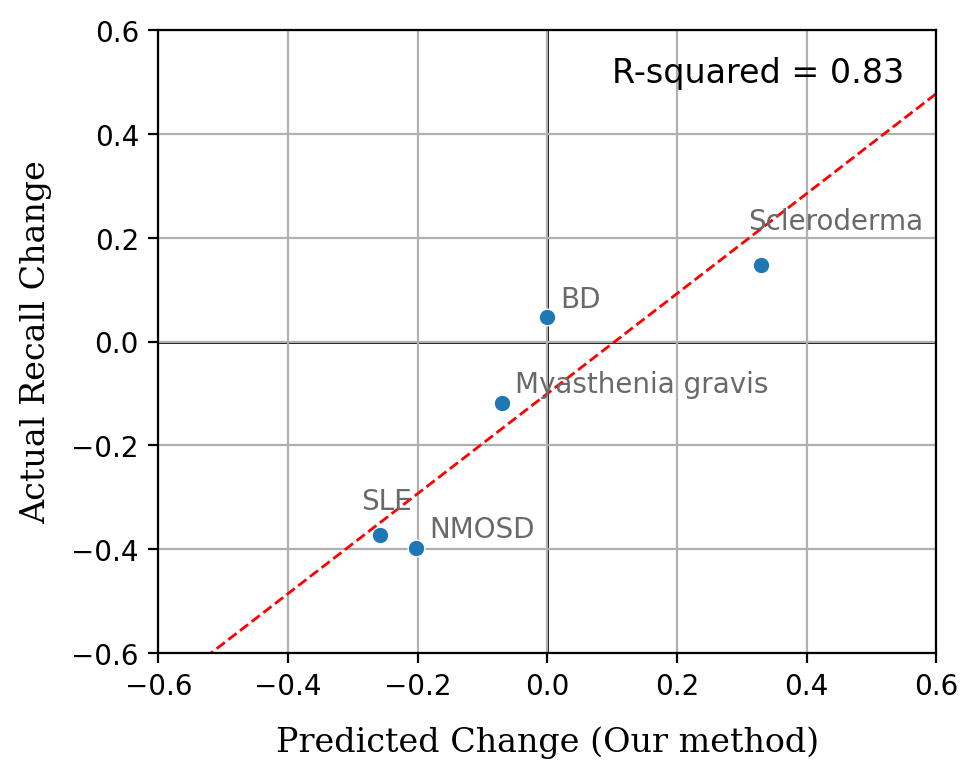}
        \captionsetup{aboveskip=4pt}
        \caption{Recall@8}
        \label{fig:lm_recall}
    \end{subfigure}
    \hspace{12pt}   
    \begin{subfigure}[b]{0.45\linewidth} 
        \centering
        \includegraphics[width=\linewidth]{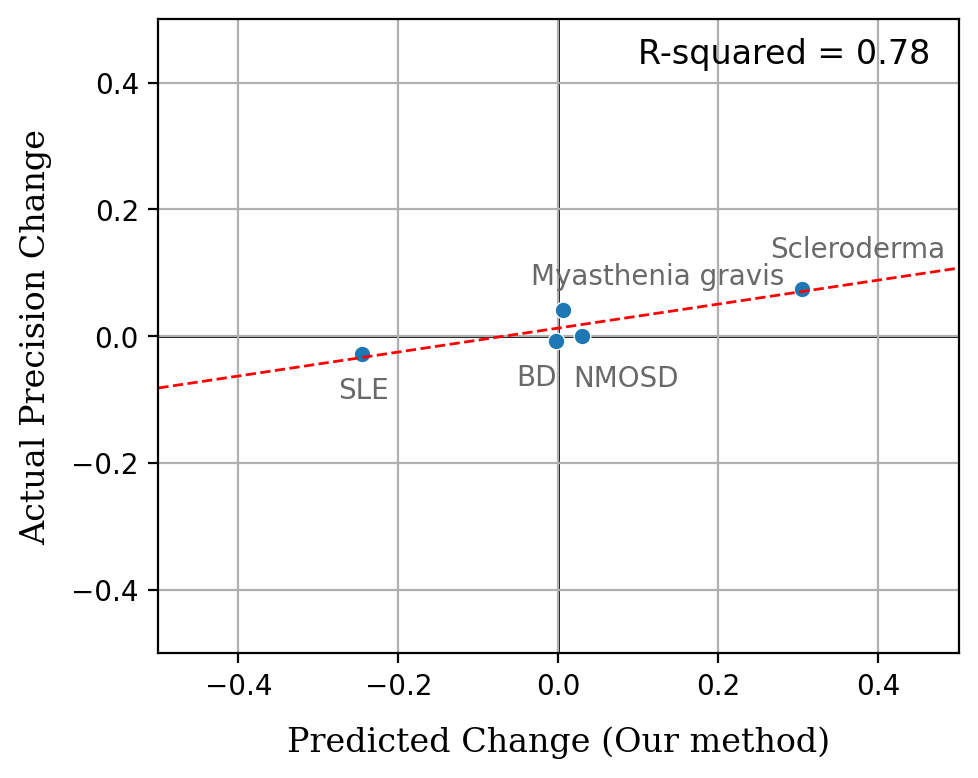}
        \captionsetup{aboveskip=4pt}
        \caption{Precision@8}
        \label{fig:lm_precision}
    \end{subfigure}
    \caption{Correlation between predicted and actual change in diagnostic performance.\\Disease with frequency annotation.}
    \label{fig:plot_with_lm}
\end{figure*}

Figure \ref{fig:lm_recall} illustrates a regression analysis of the predicted and actual recall changes for the diseases (n=5) for which frequency information is available. A high R-squared value indicates that the estimated value can accurately predict the actual value. The R-squared value between the predicted and actual values was 0.83. Also, Figure \ref{fig:lm_precision} shows the results of precision changes. The R-squared value between the predicted and actual values was 0.78.

\begin{table*}[ht]
  \captionsetup{font=large}
  \caption{Comparison of synthetic vignette simulation vs. actual changes in diagnostic performance \\for each SC$\rm{^a}$ algorithm update.}
  \begin{tabularx}{\linewidth}{lccccccccc}
    \toprule
        &
        & \multicolumn{4}{c}{Recall@8 Change [95\%CI]}
        & \multicolumn{4}{c}{Precision@8 Change [95\%CI]} \\
        \cmidrule(r){3-6} \cmidrule(r){7-10}
        Target Disease
        & Fq$\rm{^b}$
        & \multicolumn{2}{c}{\begin{tabular}{c}Synthetic Vignette \\ Simulation \\ (our method)\end{tabular}}
        & \multicolumn{2}{c}{Actual$\rm{^c}$} 
        & \multicolumn{2}{c}{\begin{tabular}{c}Synthetic Vignette\\ Simulation \\ (our method)\end{tabular}}
        & \multicolumn{2}{c}{Actual$\rm{^c}$} \\
        \midrule
        BD$\rm{^d}$ & \checkmark & \phantom{-}0.000 & [-0.031, \phantom{-}0.031] & \phantom{-}0.047 & [-0.064, \phantom{-}0.158] & -0.003 & [-0.030, \phantom{-}0.024] & -0.007 & [-0.025, \phantom{-}0.011] \\
        NMOSD$\rm{^e}$ & \checkmark &  -0.202 & [-0.342, -0.062] & -0.397 & [-0.708, -0.086] & \phantom{-}0.029 & [\phantom{-}0.017, \phantom{-}0.041] & \phantom{-}0.000 & [-0.012, \phantom{-}0.012] \\
        Scleroderma & \checkmark & \phantom{-}0.330 & [\phantom{-}0.224, \phantom{-}0.436] & \phantom{-}0.147 & [-0.006, \phantom{-}0.300] & \phantom{-}0.304 & [\phantom{-}0.284, \phantom{-}0.324] & \phantom{-}0.075 & [\phantom{-}0.055, \phantom{-}0.095] \\
        Myasthenia gravis & \checkmark & -0.070 & [-0.185, \phantom{-}0.045] & -0.118 & [-0.242, \phantom{-}0.006] & \phantom{-}0.006 & [-0.065, \phantom{-}0.077] & \phantom{-}0.042 & [-0.029, \phantom{-}0.113] \\
        SLE$\rm{^f}$ & \checkmark & -0.258 & [-0.397, -0.119] & -0.373 & [-0.642, -0.104] & -0.245 & [-0.268, -0.222] & -0.029 & [-0.052, -0.006] \\
        Multiple sclerosis & \ding{55} & -0.060 & [-0.139, \phantom{-}0.019] & \phantom{-}0.191 & [\phantom{-}0.062, \phantom{-}0.320] & -0.075 &[-0.086, -0.064] & -0.003 & [-0.014, \phantom{-}0.008] \\
        Crohn's disease & \ding{55} & \phantom{-}0.319 & [\phantom{-}0.196, \phantom{-}0.442] & \phantom{-}0.017 & [-0.120, \phantom{-}0.154] & \phantom{-}0.069 & [\phantom{-}0.065, \phantom{-}0.073] & -0.003 & [-0.007, \phantom{-}0.001] \\
        PD$\rm{^g}$ & \ding{55} & \phantom{-}0.561 & [\phantom{-}0.455, \phantom{-}0.667] & \phantom{-}0.461 & [\phantom{-}0.373, \phantom{-}0.549] & \phantom{-}0.045 & [-0.089, \phantom{-}0.179] & -0.126 & [-0.260, \phantom{-}0.008] \\
    \bottomrule
  \end{tabularx}
  \label{table:mainresult}

  \vspace{1pt}
  \begin{minipage}{\linewidth}
    \begin{flushleft}
      $\rm{^a}SC$: Symptom Checker. \\
      $\rm{^b}Fq$: Whether or not the frequency is set in the HPO annotation for the target disease. \\
      $\rm{^c}Actual$: Actual performance changes of target diseases due to the algorithm update. \\
      $\rm{^d}BD$: Behcet's disease. \\
      $\rm{^e}NMOSD$: Neuromyelitis optica spectrum disorder. \\
      $\rm{^f}SLE$: Systemic lupus erythematosus. \\
      $\rm{^g}PD$: Parkinson's disease. \\
    \end{flushleft}
  \end{minipage}

\end{table*}

Table \ref{table:summary} summarizes the estimation performance of the proposed method with $p$-values.
For all algorithm changes included in the experiment (n=8), the R-squared coefficient for Recall@8 change was 0.69 ($p$-value: 0.010), and for Precision@8 change, it was 0.16 ($p$-value: 0.318).
For algorithm updates targeting diseases with frequency information in HPO annotations (n=5),  the R-squared coefficient for Recall@8 change was 0.831 ($p$-value: 0.031), and for Precision@8 change, it was 0.78 ($p$-value: 0.047). 

We took approximately 2 hours per disease to map the Phenotypes in HPO Annotation to the SC symptoms.

\begin{table*}[ht!]
  \centering
  \captionsetup{font=large}
  \caption{Performance summary of our proposed method.}
  \begin{tabularx}{\linewidth}{lcccc}
        \toprule
        & \multicolumn{2}{c}{Recall@8 Change}
        & \multicolumn{2}{c}{Precision@8 Change} \\
        \cmidrule(r){2-3} \cmidrule(r){4-5}
        & $R^2$ & $p$-value & $R^2$ & $p$-value \\
        \midrule
        \begin{tabular}{l}All algorithm updates included in the experiment. (n=8) \vspace{3pt} \end{tabular} & 0.69 & 0.010 & 0.16 & 0.318 \\
        \begin{tabular}{l}Algorithm updates in which frequency information was included\\ in the HPO annotations for the target disease. (n=5)\end{tabular} & 0.83 & 0.031 & 0.78 & 0.047 \\
    \bottomrule
  \end{tabularx}
  \label{table:summary}
\end{table*}

\section{Discussion}
\subsection{Principal Results}
Our study found two key insights. First, for diseases with frequency information in HPO Annotations, our proposed method accurately estimates the impact of algorithm updates on individual diseases. Second, this method allows performance evaluation of rare diseases at a lower cost compared to clinical vignettes created by physicians.

Our method can predict the impact of algorithm update for diseases with frequency information in HPO Annotation. The R-squared coefficient between the predicted and actual changes for the performance metrics of algorithm update for those diseases was approximately 0.8.
Figure \ref{fig:plot_with_lm} demonstrates that actual values can be estimated through regression analysis based on our method's output. Precision (Positive Predictive Value) is affected by the prevalence rates of the dataset, so the proposed method outputs a more significant value. However, regression analysis shows that the actual values can be estimated. Conversely, the predictions tended to be inaccurate for diseases without frequency information in the HPO annotations.
Figure \ref{fig:plot_all} shows two data points that not only made a significant error in their prediction but also had the opposite of the actual change, clearly showing a prediction error. One such error was an example where a decrease in performance was predicted, but the actual performance improved.
We set all phenotypes to 50\% when there was no frequency setting, which is thought to lead to low-quality synthetic vignettes.
Reasonable estimations allow symptom checker developers to evaluate the impact of algorithm updates before release.
This enables them to avoid unexpected performance degradation, protecting users.
Furthermore, developers can efficiently improve algorithms even for rare diseases.

Our method allows performance evaluation of rare diseases at a lower cost than clinical vignettes created by physicians. In this study, a software developer, not a physician, mapped HPO phenotypes to SC symptoms. Finally, physicians reviewed the mapping.
Each disease mapping took approximately two hours.
Compared to the development of clinical vignettes, which are typically developed by a team of multiple experienced physicians [\cite{St-Marie2021-gq,Hirosawa2023-fo}]. Our method uses a database of medical knowledge that has already been reviewed so that the mapping work can be carried out by someone other than a doctor, and the final doctor's review time can be greatly reduced.
Furthermore, the proposed algorithm accepts the number of generated cases as a parameter. It probabilistically generates simulation data based on HPO Annotations, so there is no upper limit on the number.
This significantly reduces the physician's workload, enabling performance evaluations for a broader range of rare diseases.

The difference between the prevalence in the simulation dataset and the prevalence in the real world causes a difference between the output of our method and the actual scale of change in precision. 
The prevalence of the eight rare diseases in the simulation dataset was about 0.5\%, which is significantly higher than in the real world.
If the prevalence of rare diseases in the simulated dataset were to be made consistent with the real world, this difference would be reduced. However, in that case, the number of cases in the dataset would be huge.

In this study, SC symptoms mapped to HPO phenotypes were assumed to occur whenever the phenotype was present. However, the probability of occurrence should vary by symptom. For example, Table \ref{table:mappings} shows that hyperthyroidism maps to symptoms such as heat intolerance, tachycardia, and fatigue, but not all of these symptoms necessarily manifest during hyperthyroidism. Therefore, setting occurrence probabilities for each mapped symptom might enhance the effectiveness of synthetic vignettes, but the mapping cost will more than double. Also, relationships between phenotypes, such as edema being more likely in cases of renal failure, are not defined in HPO Annotations and are treated independently.

\subsection{Comparison with Prior Work}

While some studies construct performance evaluation datasets based on case reports [\cite{Miyachi2023-sr}], case reports often cover rare instances and do not adequately represent typical cases. Furthermore, extracting symptoms from case reports requires a natural language processing system. Since case reports of common diseases are scarce, our method utilizes common disease data derived from SC's user feedback.
The case of common diseases allows us to evaluate whether changes to predictive algorithms for rare diseases negatively impact the performance of common diseases.

\subsection{Limitations}
First, we evaluated the performance of our method with only a single SC. There is limited algorithm update data for rare diseases that do not have frequency information in HPO annotation.
Further validation in other SCs is necessary to establish its broader applicability and effectiveness.

Second, the value of the actual metric is difficult to predict because of the differences between the synthetic vignettes and the SC users' answers.
This is because SC input is limited to the user's self-reported symptoms. Even for symptoms that can be recognized, there is a possibility of differences with HPO Annotation for symptoms that are difficult for patients to recognize, such as jaundice.

\subsection{Conclusions}
Our simulation using medical knowledge databases showed that it is possible to estimate the impact of SC algorithm changes targeting specific diseases. This output allows SC developers to evaluate changes in predictive performance for rare diseases without needing clinician-created vignettes. 
Even though the evaluation dataset is constructed at a lower cost than clinical vignettes, it is still based on a medical knowledge database, so it is still explanatory and transparent.
Consequently, SC developers can safely update their services following pre-release testing; for example, SC developers avoid releasing changes that cause unexpected performance degradation.
This testing enables more efficient improvements in predictive performance for rare diseases, enhancing the overall quality of symptom checkers.

\end{multicols}
\section{Acknowledgements}
TN was responsible for the study design, methodology, data analysis, interpretation of the data, and writing of the manuscript. SK reviewed the manuscript and validated the data mapping. KY was responsible for conceptualizing the study and project administration. All authors reviewed the results and approved the final version of the manuscript.

\section{Conflicts of Interest}
All authors, Takashi Nishibayashi, Seiji Kanazawa, and Kumpei Yamada, are employees of Ubie, Inc.

\section{Abbreviations}
\begin{itemize}
\item SC: Symptom Checker
\item HPO: Human Phenotype Ontology
\item NMOSD: Neuromyelitis optica spectrum disorder
\item SLE: Systemic lupus erythematosus
\item PD: Parkinson's disease
\item BD: Behcet's disease
\item PD: Parkinson's disease
\end{itemize}
        

\printbibliography

\end{document}